\documentclass{article}

\usepackage[preprint]{corl_2026} % Use this for the initial submission.
\usepackage{amsmath}
\usepackage{amssymb}
\usepackage{booktabs}
\usepackage{newunicodechar}
\usepackage{graphicx}
\usepackage{float}
\newunicodechar{δ}{\delta}

\title{What Do They See? Interpreting Complex Road Scenarios Through the Eyes of Vision-Language-Action Models for Safe and Trustworthy Autonomous Vehicle Learning}

\author{
  \begin{tabular*}{\textwidth}{@{\extracolsep{\fill}}cccc@{}}
    Kalpana Panda\textsuperscript{1,2} &
    Wesley Maia\textsuperscript{2} &
    Vinti Agarwal\textsuperscript{1} &
    Ross Greer\textsuperscript{2}
  \end{tabular*}
  \\[1em]
  \textsuperscript{1}Department of Computer Science and Information Systems, \\
  Birla Institute of Technology and Science, Pilani, India \\
  \textsuperscript{2}Machine Intelligence, Interaction, and Imagination (Mi$^3$) Lab, \\
  University of California, Merced, United States
}

\begin{document}
\maketitle

%===============================================================================

\begin{abstract}
    End-to-end autonomous driving models are now able to navigate complex road scenarios, mapping raw sensor observations directly to observed paths for open-loop evaluation and often effective driving in closed-loop evaluation. Yet the internal logic of these safety-critical systems remains largely opaque, due to the complexity of traffic scenes. We propose a counterfactual ablation framework called Counterfactual Vision Action Analysis (CVAA) that systematically removes individual detected objects from front-camera images using photorealistic generative inpainting to prepare counterfactual sets to evaluate the difference in the model’s response.  This isolates the causal effect of each object's presence on the model's planning behaviour. Applied to the Alpamayo 1 trajectory predictor across 210 nuScenes driving scenes, we create a dataset Counter -nuScenes, using which we see that vehicles and pedestrians within the model’s ‘path’ dominate causal influence as expected, while traffic lights, as expected, exert disproportionate effect relative to their image footprint. However, we also find cases where the model responds strongly to objects a human driver would consider irrelevant. 
This brings forth a deeper question: does the model itself view the scene as a sum of individual objects influencing the outcome, or does it encode an entirely different set of internal features that do not correspond to human-legible scene elements? To further understand this, we compare intermediate representations of original and inpainted image pairs using mechanistic interpretability techniques and examine the effect of the removal through the various model layers. Together, these two stages offer a path from behavioral auditing to representational understanding, creating explainable driving systems and solidifying human-AI trust.

\end{abstract}


%===============================================================================

\section{Introduction}
\label{sec:introduction}

Autonomous driving systems must interpret complex scenes containing interacting vehicles, pedestrians, cyclists, traffic signals, road infrastructure, and work-zone elements. Large-scale datasets such as KITTI~\citep{Geiger2012KITTI}, nuScenes~\citep{Caesar2020nuScenes}, the Waymo Open Dataset~\citep{Sun2020Waymo}, and BDD100K~\citep{Yu2020BDD100K} have driven major progress in perception, tracking, forecasting, and planning. More recently, autonomous-driving research has moved toward unified models that connect visual scene understanding with language reasoning and action generation, including UniAD~\citep{Hu2023UniAD}, DriveGPT4~\citep{Xu2023DriveGPT4}, LMDrive~\citep{Shao2024LMDrive}, DriveLM~\citep{Sima2024DriveLM}, DriveVLM~\citep{Tian2024DriveVLM}, and EMMA~\citep{Waymo2024EMMA}.

Despite this progress, it remains difficult to determine which scene objects actually influence a model's driving decision. A model may generate a plausible trajectory or a convincing language rationale without being causally driven by the correct visual evidence. This is especially concerning in safety-critical settings: a driving model should respond to a lead vehicle, crossing pedestrian, or traffic light, while remaining relatively insensitive to irrelevant background objects. We therefore ask an object-level causal question: which visible objects drive a vision-based model's trajectory prediction?

Existing causal and perturbation-based evaluations in driving typically operate on structured state representations, simulator abstractions, or agent trajectories~\citep{DeHaan2019CausalConfusion, Bansal2019ChauffeurNet, Roelofs2023CausalAgents}. Visual attribution methods such as saliency maps, Grad-CAM, RISE, and SHAP~\citep{Simonyan2014Saliency, Selvaraju2017GradCAM, Petsiuk2018RISE, Lundberg2017SHAP} instead operate on images, but pixel-level explanations can be hard to audit in dense traffic scenes and may conflate object importance with artifacts introduced by position, masking, or occlusion \cite{greer2022salience, greer2023robust, greer2023salient}. We address this gap by intervening directly on semantic objects in the camera image while keeping the edited scene visually plausible.

We introduce \emph{Counterfactual Vision Action Analysis (CVAA)}, a counterfactual object-attribution framework, and \emph{Counter-nuScenes}, a benchmark dataset for vision-based autonomous-driving models. Given a front-camera traffic scene, we detect and segment individual objects, remove each one using photorealistic generative inpainting, rerun trajectory inference, and measure the resulting shift relative to the model's original trajectory distribution. By replacing removed objects with contextually plausible background rather than black boxes, blur, or simple occlusion, the framework aims to isolate the causal effect of each object's presence while reducing distribution shift from unrealistic perturbations~\citep{Chang2019ExplainingCounterfactuals, Zhang2020AutoRemover, lama,ravi2024sam2}.

We evaluate this framework on nuScenes front-camera scenes using the Alpamayo 1 trajectory predictor. For each scene, we construct a counterfactual semantic set in which each detected object is removed independently, and we rank objects by trajectory-shift metrics computed against the model's own original prediction. Unlike minADE and minFDE, which measure accuracy against a singular ``ground truth" instantiation of all possible driving behaviors \cite{greer2021trajectory}, our metrics measure behavioral sensitivity: the goal is not to evaluate whether the model predicts the human driver correctly, but to identify which objects causally affect its output. Our trajectory-aimed approach removes the additional layer of oversimplification of a text output given, which may not register changes in trajectory as long as the broad path of the vehicle remains the same. This helps to further enhance the safety, as even slight trajectory changes can prove to be injurious in certain scenes. This is especially important in the context of safety, as it is not the verbal explanation which determines the safety of the scene, but rather the output trajectory which dictates vehicle motion.

Beyond black-box behavioral attribution, we also use the resulting high-influence objects as entry points for white-box analysis. Comparing original and inpainted representations, and applying mechanistic interpretability tools such as sparse autoencoders and activation patching~\citep{Elhage2021TransformerCircuits, Meng2022ROME, Bricken2023Monosemanticity, Cunningham2023SAE}, we analyze whether behaviorally important objects correspond to recoverable internal features or causal model components.

\textbf{Contributions.}
\textbf{(C1)} We introduce \textbf{Counter-nuScenes}, a counterfactual benchmark built from $210$ nuScenes scenes with photorealistic per-object inpainting, producing $3{,}062$ in-distribution counterfactual image pairs.
\textbf{(C2)} We propose \textbf{CVAA AD} and \textbf{CVAA FD}, two seed-stable distributional shift metrics that quantify each object's causal influence on predicted trajectories without requiring ground-truth poses.
\textbf{(C3)} We provide a \textbf{black-box evaluation} of Alpamayo 1, systematically measuring attribution patterns across object class, mask size, and scene structure.
\textbf{(C4)} We conduct a \textbf{white-box mechanistic analysis} that traces layer-wise causal propagation, identifies four propagation regimes, and reveals the trajectory expert as a nonlinear amplifier of object-level perturbations.

\section{Related Work}
\label{sec:related}

\textbf{Unified driving models.}
Autonomous-driving research has shifted from modular perception--prediction--planning stacks toward models that couple scene understanding, reasoning, and action. Planning-oriented systems such as UniAD~\citep{Hu2023UniAD} optimize intermediate perception and forecasting for downstream planning, while driving VLM/VLA systems such as DriveGPT4~\citep{Xu2023DriveGPT4}, LMDrive~\citep{Shao2024LMDrive}, DriveLM~\citep{Sima2024DriveLM}, DriveVLM~\citep{Tian2024DriveVLM}, OpenDriveVLA~\citep{Zhou2026OpenDriveVLA}, AutoVLA~\citep{AutoVLA2025}, Reasoning-VLA~\citep{ReasoningVLA2025}, and Alpamayo 1~\citep{AlpamayoR1,nvidiaAlpamayo} combine visual context with language or action generation. These systems improve inspectability, but generated rationales do not by themselves establish which visual entities causally influenced the final trajectory.

\textbf{Causal evaluation and visual attribution.}
Causal confusion, ChauffeurNet, and CausalAgents show that learned driving policies and forecasters can depend on non-causal correlates or can be evaluated through targeted perturbations~\citep{DeHaan2019CausalConfusion, Bansal2019ChauffeurNet, Roelofs2023CausalAgents}. In parallel, saliency, Grad-CAM, meaningful perturbations, RISE, and SHAP provide image-level attribution tools~\citep{Simonyan2014Saliency, Selvaraju2017GradCAM, Fong2017MeaningfulPerturbation, Petsiuk2018RISE, Lundberg2017SHAP}. Object-level methods such as PixelSHAP move closer to traffic-scene auditing~\citep{Goldshmidt2025PixelSHAP}, but artificial masks or occlusions can introduce out-of-distribution evidence. Our work combines the causal question from driving perturbation studies with object-level interventions performed directly on camera images.

\textbf{Counterfactual editing and mechanistic analysis.}
Generative counterfactual explanations and inpainting-based editing reduce perturbation artifacts by replacing removed content with plausible background~\citep{Chang2019ExplainingCounterfactuals, Zhang2020AutoRemover}. Modern tools such as SAM2~\citep{ravi2024sam2} and LaMa~\citep{lama} make object-level counterfactual image sets practical at scale. Behavioral attribution, however, still leaves open how influential objects are represented internally. Mechanistic interpretability methods, including causal tracing, activation patching, and sparse autoencoders~\citep{Elhage2021TransformerCircuits, Meng2022ROME, Bricken2023Monosemanticity, Cunningham2023SAE}, and recent multimodal analyses such as NOTICE and VLM-SAE work~\citep{Golovanevsky2025NOTICE, Pach2025VLM_SAE}, provide tools for connecting object-level counterfactual effects to internal model features.

\section{Methodology}
\label{sec:methodology}

Our framework, \emph{Counterfactual Vision Action Analysis (CVAA)}, probes the causal influence of individual scene objects on the trajectory predictions of a VLA-based autonomous driving model by constructing a controlled counterfactual benchmark and measuring the resulting distributional shift in predicted futures. We describe each component in turn.

\subsection{Benchmark Construction}
\label{sec:benchmark}

\textbf{Base dataset.}
Our Counter-nuScenes benchmark is built on the nuScenes v1.0-trainval split~\cite{Caesar2020nuScenes}, which comprises 700 driving scenes of approximately 20 seconds each, recorded in Boston and Singapore across a range of urban traffic conditions. Each scene provides six temporally synchronized camera streams, LiDAR sweeps, and high-frequency ego-pose measurements from GPS/IMU at up to 20\, Hz.

\textbf{Frame selection.}
We restrict our evaluation to front-camera (\texttt{CAM\_FRONT}) keyframes, which focus on the model's primary visual input and support a more localized interpretation of counterfactual effects. To ensure sufficient ego-motion context, we retain only frames preceded by at least 1.6 seconds of continuous pose data, matching the history window expected by the model. Within each scene, we detect the number of visible agents using YOLOv8~\cite{Jocher2023YOLOv8} and select the single keyframe with the highest object count, yielding one maximally informative observation per scene. From the resulting pool, after cleaning the data, we sample \textbf{210 frames}, spanning scenes with 6 to 32 valid detected objects and covering a wide range of traffic densities.

\textbf{Object segmentation.}
For each selected frame, we apply the Segment Anything Model 2 (SAM2)~\cite{ravi2024sam2} to obtain per-object instance masks. SAM2 is prompted with the bounding boxes produced by YOLOv8, and its masks are refined at the pixel level to avoid bleeding across object boundaries.

\subsection{Counterfactual Semantic Sets}
\label{sec:counterfactuals}

For the purpose of CVAA and given the instance masks for a scene, we construct a \emph{counterfactual semantic set Counter-nuScenes}: a collection of images in which each object has been individually removed via photorealistic inpainting while all other scene content remains unchanged. This produces a paired corpus of one original image and $N$ counterfactual images per scene, where $N$ is the number of detected objects.
\begin{figure}[h]
    \centering
    \includegraphics[width=0.98\linewidth]{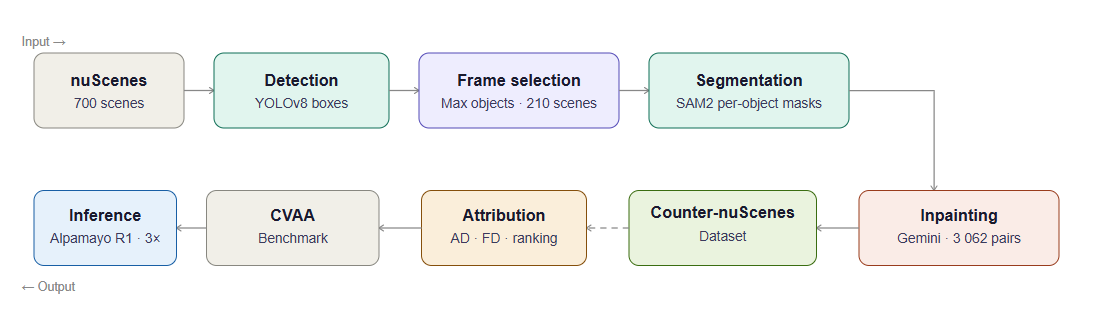}
   \small\caption{Counter-nuScenes and CVAA construction pipeline.}
    \label{fig:pipeline}
\end{figure}
\textbf{Inpainting approach.}
Object removal is performed using Gemini's \textit{imgen} inpainting tool~\cite{gemini2024}. We also maintain an open-source ablation pipeline based on LaMa~\cite{lama} combined with FLUX Fill 1.0~\cite{flux2024}. We deliberately use photorealistic inpainting rather than simpler alternatives such as color filling, Gaussian blurring, or copy-paste occlusion. This design choice distinguishes our setting from prior attribution methods in autonomous driving, where masks were blacked out~\cite{Goldshmidt2025PixelSHAP} or semantic image pairs were used as surrogates~\cite{Golovanevsky2025NOTICE}. Photorealistic inpainting keeps counterfactual images closer to the model's data distribution: the intended change is confined to the masked region and corresponds to the semantic absence of the object, rather than to distribution shift introduced by an unrealistic perturbation. Consequently, changes in model output can be interpreted as responses to the removed object rather than to rendering artifacts.

\subsection{Attribution Metrics}
\label{sec:metrics}

For each scene, we run the model on the original image to obtain a reference trajectory distribution and on each counterfactual image to obtain a variant distribution. We quantify the causal influence of each removed object by measuring the displacement between these two distributions.

\textbf{Trajectory representation.}
The model produces $K$ stochastic trajectory samples $\{\mathbf{p}_k\}_{k=1}^K$, where each $\mathbf{p}_k$ is a sequence of $T$ future ego-centric waypoints. We project trajectories onto the XY plane and summarize each predictive distribution by its mean trajectory, which is more stable across random seeds than selecting a single sample.

\textbf{Average Deviation (AD) and Final Deviation (FD).}
Let $\bar{\mathbf{p}}^{\mathrm{orig}}$ and $\bar{\mathbf{p}}^{\mathrm{var}}$ denote the mean trajectories of the original and counterfactual runs, respectively. We define:
\begin{equation}
\begin{aligned}
\bar{\mathbf{p}} &= \frac{1}{K}\sum_{k=1}^{K}\mathbf{p}_k \in \mathbb{R}^{T\times 2},\\
\mathrm{AD} &= \frac{1}{T}\sum_{t=1}^{T}\left\|\bar{\mathbf{p}}^{\mathrm{var}}_t-\bar{\mathbf{p}}^{\mathrm{orig}}_t\right\|_2,\quad
\mathrm{FD} = \left\|\bar{\mathbf{p}}^{\mathrm{var}}_T-\bar{\mathbf{p}}^{\mathrm{orig}}_T\right\|_2.
\end{aligned}
\label{eq:attr_metrics}
\end{equation}

AD captures the \emph{integrated} influence of an object over the full prediction horizon, while FD isolates its effect on the \emph{terminal position}, a quantity directly relevant to planning safety margins. Together, they provide complementary views: an object may have a large AD, indicating early influence that later decays, but a modest FD, or vice versa.

\textbf{Object-level attribution score.}
For a scene with $N$ objects, removing object $i$ yields the pair $(\mathrm{AD}_i,\, \mathrm{FD}_i)$. We rank objects by $\mathrm{AD}$ as our primary attribution score, breaking ties by $\mathrm{FD}$. This ranking constitutes our \emph{counterfactual object attribution} (COA) ordering for the scene. Aggregating COA rankings across all 210 scenes allows us to analyze which object categories are systematically most influential to the model's trajectory predictions, and to compare the structure of these attributions between the black-box and white-box evaluation regimes (Section~\ref{sec:experiments}).

\textbf{Relationship to existing metrics.}
AD and FD are conceptually related to the standard minADE and minFDE metrics used in trajectory forecasting~\cite{Caesar2020nuScenes}, but serve a fundamentally different purpose. minADE and minFDE measure \emph{accuracy} relative to a ground-truth future, selecting the best candidate trajectory. AD and FD instead measure \emph{distributional shift} relative to a reference prediction, using the full candidate set. Using a min-over-$K$ estimator here would be inappropriate: it would reward any variant that happens to produce one trajectory close to the original distribution by chance, conflating sampling noise with causal influence. The mean-based estimator in Equation~\ref{eq:attr_metrics} avoids this confound.

\section{Experimental Setup}
\label{sec:experiments}

To investigate Alpamayo's behavior, we evaluate the model under two complementary regimes. First, in the \textbf{black-box} setting, we treat Alpamayo 1 as an opaque function and measure object influence solely through AD and FD on \emph{CVAA} (Section~\ref{sec:metrics}). We run Alpamayo 1 three times on the dataset using seeds 42, 7, and 123. During these runs, inputs from all cameras except the front camera are zeroed out, and the model receives a single still image for each original or inpainted scenario. This isolates the effect of removing one object at a time from the front-camera view. We also create a dashboard \ref{app:scene_explorer} that gives a visual reference to compare the importance of different objects in a scene.

Second, in the \textbf{white-box} setting, we use the  \emph{Counter-nuScenes} image pairs to inspect internal model behavior. Alpamayo 1 comprises a vision encoder (27 blocks, a VLM (36 layers,
3006-token sequence with 180 tokens per camera image after spatial
merging), and a trajectory expert (36 denoising layers).
 
For each removed object, its bounding box is projected onto the
model's patch grids to identify the corresponding token positions
$\mathcal{T}$ in both the vision encoder and VLM.
We compute layer-wise cosine deltas between original and inpainted
hidden states at object, visual, and global token positions across
all three modules~\cite{Palit_2023_ICCV, haon2025,neoICLR2025}, yielding a per-layer
attribution signal that localises where in the model the object's
removal registers.
We additionally apply a Logit Lens~\cite{nostalgebraist2020} to
track how the object's class representation evolves across layers,
compute spatial KL divergence maps over visual patch positions~\cite{neoICLR2025}, and extract attention weights from the trajectory handoff token to the object's patch positions.
Delta curves are computed for all layers; Logit Lens and attention
analyses are run at representative layers in each module.
% ─────────────────────────────────────────────────────────────────────────────
\section{Results}
\label{sec:results}
\subsection{Black-Box Evaluation}
\label{sec:blackbox}
 
\textbf{Per-Class Deviation.}
 Table~\ref{tab:class_deviation} reports mean AD and FD per object class.
\texttt{Class bus} seems to produce the highest deviation ($\text{AD} = X, Y$\,m), while \texttt{class umbrella} objects appear to have negligible impact.
However, this is additionally affected by their lower numbers, resulting in relatively skewed AD/FD statistics, while more popular object classes have their influence more smoothed out. One notable point, however, is the influence that traffic lights appear to possess, outranking both cars and people, which is analogous to their importance on the road. 

\textbf{Rank Stability.}
 Since raw AD and FD statistics are scene-specific, we proceed to evaluate objects on the basis of their ranking within the scenes, based on AD and FD. On checking the rank stability between runs with different scenes, an interesting point to note is that objects rarely maintain their rank. As visible in ~\ref{tab:rank_stability}, only 5.6\% of objects keep their rank with respect to AD, and 6.4\% with respect to FD, with 14.6\% and 16\% of objects in a top 3 rank remaining within the top 3, respectively, with the average rank range being 4.691 for AD and 4.641 for FD. This strongly indicates that object importance is not stable and hence that the model rarely takes cues from a single object. 

\begin{table}[t]
  \centering
  \scriptsize % Noticeably decreases font size to look tight and crisp
  \caption{\textbf{Rank Stability Summaries (3 Seeds).}}
  \label{tab:rank_stability}
  
  % Left Table: AD Metric
  \begin{minipage}{0.48\linewidth}
    \centering
    \begin{tabular}{|l|c|}
      \hline
      \multicolumn{2}{|c|}{\textbf{Rank Stability Summary (AD)}} \\
      \hline
      \textbf{Top-3 Objects} & \\
      ~~Exact rank kept & 14.6\% \\
      ~~Remain in top-3 & 39.7\% \\
      \hline
      \textbf{All Objects} & \\
      ~~Exact rank kept & 5.7\% \\
      ~~Rank shift $\le \pm 2$ & 34.1\% \\
      \hline
    \end{tabular}
  \end{minipage}
  \hfill
  % Right Table: FD Metric
  \begin{minipage}{0.48\linewidth}
    \centering
    \begin{tabular}{|l|c|}
      \hline
      \multicolumn{2}{|c|}{\textbf{Rank Stability Summary (FD)}} \\
      \hline
      \textbf{Top-3 Objects} & \\
      ~~Exact rank kept & 16.0\% \\
      ~~Remain in top-3 & 39.5\% \\
      \hline
      \textbf{All Objects} & \\
      ~~Exact rank kept & 6.4\% \\
      ~~Rank shift $\le \pm 2$ & 35.2\% \\
      \hline
    \end{tabular}
  \end{minipage}
\end{table}

\textbf{Scene-Level Deviation.}
Based on rankings derived by averaging AD and FD for each object across the three runs, we also examine scenes individually.  

\textbf{Per-scene distribution of importance.}
Figure~\ref{fig:zscore} shows the per-scene z-score of the
maximum-AD object, where $z = (\max(\text{AD}) - \bar{\text{AD}}) /
\sigma_{\text{AD}}$ measures how many standard deviations the
most influential object lies above the scene mean.
The distribution spans $z \in [1.0,\ 4.52]$ with mean $2.43$,
median $2.35$, and IQR $[1.91,\ 2.96]$, indicating that in the
typical scene some object exerts a disproportionate influence
relative to its peers.
148 of 210 scenes (70.5\%) exceed the conventional $z{>}2$ threshold,
and 48 scenes (22.9\%) exceed $z{>}3$.
FD z-scores are closely aligned (mean $2.41$, median $2.34$,
$z{>}2$: 148, $z{>}3$: 43), with no systematic decoupling between
integrated and terminal deviation.
 
The right-skewed tail reflects the structural heterogeneity of urban
driving scenarios: dense junction scenes distribute influence across
many agents and suppress $z$, while sparse environments (where a
single isolated object is the dominant scene cue) drive $z$ into
the extreme tail.
This dominance is best understood as a property of the scene's
deviation landscape rather than a stable attribution to one specific
object, as rank orderings show considerable seed-to-seed variance
(Table~\ref{tab:rank_stability}). This suggests that the model is sensitive to the presence of a high-influence region rather than to one uniquely identifiable agent, which we discuss further in Section~\ref{sec:conclusion}
.
\textbf{Object count vs peak deviation per scene.}
Crucially, object count shows \emph{no relationship} with peak deviation (Spearman $\rho =0.11$ for AD,+ $0.12$ for FD), suggesting that object presence in certain high-priority regions matters more than scene density in driving trajectory shifts.

\textbf{Mask Size.}
Mask area correlates positively with both AD and FD (Pearson $r = +0.25$, Spearman $\rho = +0.22$), indicating that physically larger objects exert greater causal influence on the
model's predicted trajectory when removed.
This is further reflected in the rank structure: rank-1 objects (highest AD) average 38{,}706 pixels, falling sharply to 13{,}932 at rank 2 and below 5{,}000 beyond rank 5 (Spearman $\rho = -0.56$, $p \ll 0.001$), suggesting that object scale is a primary structural predictor of deviation within a scene.

\textbf{Traffic Light Colour.}
Of 531 traffic light objects, pixel-level colour classification
detected 119 (22.4\%); the remainder lacked a clearly dominant
red or green channel, likely due to grey housing, night conditions,
or overexposure.
Among detected lights, red-state lights produce substantially higher
deviation than green-state (mean AD $0.758$ vs $0.229$\,m; mean FD
$2.241$ vs $0.672$\,m), consistent with the intuition that a
stationary ego at a red light has a more constrained and predictable
trajectory. Thus, removing an object in that context causes a larger
distributional shift than when the ego is in motion at a green light.

\textbf{Distributional Shift vs.\ Ground-Truth Accuracy.}
Of 2{,}975 objects, 1{,}468 (49.3\%) exhibit a negative
$\Delta\,\text{minADE}$ , meaning their removal causes the model's best
trajectory to move \emph{closer} to ground truth despite a
distributional shift.
Among the 360 high-AD objects (top 25\% by AD, threshold $0.516$\,m),
the Spearman correlation between AD and $\Delta\,\text{minADE}$ is
near zero ($\rho = +0.03$), confirming that the magnitude of
distributional shift carries no consistent signal about whether
accuracy improves or degrades.
The most extreme case is scene-0708, where two cars produce
$\Delta\,\text{minADE} = -2.49$ and $-1.90$\,m respectively,
suggesting the model was substantially over-relying on those
objects.

%%\paragraph{Lateral Position}
%Objects on the right side of the frame produce notably higher
%deviation than those on the left or centre (mean AD: right $0.530$\,m,
%centre $0.396$\,m, left $0.366$\,m; mean FD: right $1.503$\,m,
%centre $1.136$\,m, left $1.035$\,m).
%However, the continuous Spearman correlation between normalised
%x-position and AD is weak ($\rho = +0.06$), indicating that the
%right-side effect is driven by a subset of high-influence objects
%rather than a smooth positional gradient.
%This asymmetry may reflect nuScenes scene composition, right-side
%objects in urban driving often include oncoming traffic, parked cars
%at junctions, and traffic infrastructure, which are likely to be
%semantically salient to the model.

\textbf{Per-Class Rank Distribution.}
Table~\ref{tab:rank1} shows how often each class occupies the
top AD rank across 210 scenes.
Cars rank first most frequently (78 scenes, 37.1\%), followed by
persons (48, 22.9\%) and traffic lights (41, 19.5\%).
However, when normalised by class count, buses and trucks rank
first at a disproportionately high rate (19.1\% and 16.4\% of
their appearances respectively), consistent with their larger
physical scale and correspondingly larger masks.

The interesting observation, however, comes in the form of certain seemingly ranking objects having disproportionately high rankings.
In 1.0\% of scenes, a stop sign ranks above all other agents,
and in one scene a tennis racket is the single most influential
object.
Even within plausible classes, the ranking frequently defies
intuition: a pedestrian who has already crossed the road above
one stepping into the ego's path, a ball-shaped fixture on the side of the road above someone crossing the road, a traffic light whose signal isn't visible from the photograph angle over a car immediately ahead of the camera.
These cases seem to be the rule, not the exception. 
49.3\% of objects improve ground-truth accuracy when removed
 suggesting the model is not simply
tracking the most safety-critical agent, but looking at the image a lot more holistically than we may perceive. Additionally, it points to the fact that there are multiple objects that the model draws consolidated cues from (a person on the crosswalk and 2 different red light signals in front of the car all point to the fact that the vehicle should stop). Combined with the low rank stability across seeds
(Section~\ref{tab:rank_stability}), this points to a fundamental
limitation of single-object counterfactual attribution: the
model's response to any one removal is entangled with its
broader scene representation.

\begin{table}[h]
\centering
\caption{Per-class rank-1 frequency across 210 scenes. Classes never ranking first are omitted.}
\label{tab:rank1}
\scriptsize
\begin{tabular}{lrrrrr}
\toprule
\textbf{Class} & \textbf{N} & \textbf{\#1} & \textbf{\%1} & \textbf{\#Top\,3} & \textbf{\%1/Total} \\
\midrule
Car           & 1566 & 78 & 37.1 & 272 &  5.0 \\
Person        &  604 & 48 & 22.9 & 134 &  7.9 \\
Traffic light &  531 & 41 & 19.5 & 122 &  7.7 \\
Truck         &  183 & 30 & 14.3 &  61 & 16.4 \\
Bus           &   47 &  9 &  4.3 &  18 & 19.1 \\
Stop sign     &    8 &  2 &  1.0 &   3 & 25.0 \\
Tennis racket &    1 &  1 &  0.5 &   1 & 100.0 \\
Bicycle       &   28 &  1 &  0.5 &   7 &  3.6 \\
\bottomrule
\end{tabular}
\end{table}
% ─────────────────────────────────────────────────────────────────────────────
\subsection{White-Box Evaluation}
\label{sec:whitebox}

We structure our analysis layer-by-layer, tracing how the removal of
each object propagates through Alpamayo 1's three modules: vision
encoder, language model, and trajectory expert.

% ─────────────────────────────────────────────────────────────────────────────
\textbf{Visual Encoding and LM Propagation.}
\label{sec:wb_encoding} 
The vision encoder delta at the removed object's patch positions is
on the order of $10^{-7}$ and negative across all categories
(Figure~\ref{fig:delta_curves}, bottom right), indicating the
inpainted region is more self-consistent in feature space than the
original, a consequence of photorealistic inpainting.
Despite this, the signal does not transfer cleanly into the LM:
the correlation between late vision encoder delta and early LM delta
is weak or negative for all classes (Pearson $r \in [-0.41, +0.13]$),
with a sign reversal for car ($-5.94$) and other ($-4.81$), suggesting
the two representations diverge differently at the merge step. Additionally, curves are identical across categories at this stage, indicating that the vision encoder does not differentiate between object classes.
 
Within the LM, all signals are flat for layers 0--15.
From layer 15, curves diverge sharply by category: bus reaches
${\approx}0.06$ at layer 35 in the object-token panel, truck
${\approx}0.02$, while car and bicycle remain near zero throughout
(Figure~\ref{fig:delta_curves}, top left).
Comparing the object-token and all-visual delta panels shows the
perturbation is spatially contained for most classes; bus is the
exception, where the all-visual delta substantially exceeds the
object-token delta, indicating the model re-interprets surrounding
context when a large occluding vehicle is removed.
 
% ─────────────────────────────────────────────────────────────────────────────
\textbf{Trajectory Handoff and Four Propagation Regimes.}
\label{sec:wb_handoff}
The global delta (last token, Figure~\ref{fig:delta_curves}, bottom
left) peaks at layers 20--25 before decaying, with bus and truck
peaking highest.
This fall in the curve indicates that, despite recognition of an extra object in the original case, it resolves to a stabilised state (the difference between the two is suppressed). This explains the globally small handoff delta (median $\delta_{\text{traj}}=0.0002$, confirming that most object absences barely register at the VLM--expert interface.
 
Classifying variants by whether $\delta_{\text{traj}}$ and
$|\Delta\,\text{minADE}|$ are each above or below their medians
reveals four regimes.
\emph{Coupled} variants (approx. 33.0\%) propagate cleanly from LM to
output; most prevalent in bus (62.9\%) and motorcycle (66.7\%).
\emph{Transparent} variants (approx. 33.1\%) produce negligible signal at
every stage; dominant in car (36.2\%).
\emph{Decoupled} variants (approx. 16.9\%) register at the handoff,
often via a late spike through the LM, but the both seem to converge; the model noticed two different scenes acted the same despite it.
\emph{Silent} variants (approx. 16.9\%) show negligible handoff delta yet
produce substantial output AD (up to $1.18$\, in
scene-0293), indicating influence that bypasses the trajectory
token entirely.
 
% ─────────────────────────────────────────────────────────────────────────────
\textbf{Expert Amplification.}
\label{sec:wb_expert}
Across all categories and 100\% of variants, the trajectory expert
reacts in a way that causes divergence based on what it received from the VLM (mean $115$--$180\times$,
median $47$--$106\times$).
Amplification is most extreme for silent variants, where
$\delta_{\text{traj}} \approx 0.0001$ yet ratios exceed $1000\times$
in scenes 0293, 0464, and 0637.
The expert is therefore not a passive decoder: it is a sensitive
nonlinear amplifier that resolves near-zero handoff perturbations
into large distributional shifts, and is the primary mechanism by
which silent variants produce high output deviation despite leaving
no trace at the trajectory token.
% ─────────────────────────────────────────────────────────────────────────────
\section{Discussion}
\label{sec:discussion}
 
The black-box and white-box results are broadly consistent but reveal
a more complex picture than either alone would suggest.
Individual object rankings are unstable, indicating that class-level patterns are
reliable while object-level attribution is not.
The four propagation regimes show that the trajectory handoff token
is not a universal bottleneck: silent variants exert causal influence
through a pathway that bypasses it entirely, while decoupled variants
confirm the expert actively suppresses differences it receives rather
than faithfully propagating them.
The LM's late-layer convergence between the two runs, the global
delta peaking then decaying before handoff is consistent with
a strong scene prior that stabilises the representation regardless
of which specific objects are present, which may explain the model's
robustness to transparent object removals in dense scenes.
The Logit Lens drop ($P_{\text{orig}} - P_{\text{inp}}$, ${\sim}10^{-5}$)
is not only near-zero but incoherent: classes such as bus assign
\emph{higher} probability to their own label in the inpainted run
than in the original, with no consistent relationship to the object's
actual presence.
This indicates the model does not perceive object identity as a
vocabulary-level concept at any layer: the visual tokens encode
scene geometry and relational context, not categorical object labels,
and standard language model interpretability tools do not transfer
to this representational regime.
 
\section{Limitations}
\label{sec:limitations}
 
Our evaluation is limited to a single model (Alpamayo 1) and the
nuScenes front-camera distribution; generalisation to other VLAs
and datasets remains unvalidated.
More fundamentally, the white-box analysis exposes the absence of
interpretability tools designed for VLA architectures.
Cosine delta curves and Logit Lens, both adapted from language
model research, measure what changes between two forward passes
but cannot explain \emph{why}, nor identify which internal
computations are causally responsible for the output difference.
The silent quadrant in particular, where large output deviation
occurs without any signature at the trajectory token, points to
influence pathways that existing probes cannot resolve.
Developing probes that operate in the continuous geometry of
multimodal hidden states, rather than through vocabulary projection,
is a prerequisite for deeper mechanistic understanding of VLA
decision-making.
 
\section{Conclusion}
\label{sec:conclusion}
 
We introduced CVAA and Counter-nuScenes, a counterfactual model-agnostic benchmark and dataset respectively that
measures the causal influence of individual scene objects on VLA
trajectory predictions via photorealistic inpainting and
distributional deviation metrics (AD and FD).
Applied to Alpamayo 1, the pipeline surfaces consistent class-level
sensitivities: large vehicles cause the most deviation, mask size
and lateral position are structural predictors. White-box
analysis, on the other hand, reveals that this influence operates through multiple
internal pathways, not all of which are visible at the trajectory
handoff. Our work thus opens a window of insight into how autonomous vehicles 'see' scenes as more than the sum of individual objects, where no one part can heavily disproportionately impact the models' outputs.
Future work includes applying it to other VLAs, evaluating text-similarity in VLA prompted outputs, incorporating temporal
multi-frame counterfactuals, and developing interpretability probes
suited to scene-level relational representations.

\clearpage
% The acknowledgments are automatically included only in the final and preprint versions of the paper.
% \acknowledgments{If a paper is accepted, the final camera-ready version will (and probably should) include acknowledgments. All acknowledgments go at the end of the paper, including thanks to reviewers who gave useful comments, to colleagues who contributed to the ideas, and to funding agencies and corporate sponsors that provided financial support.}

%===============================================================================

\bibliography{example}  % .bib
\appendix

\section{Supplementary Tables and Figures}
\label{app:figures}

This appendix contains supporting visualisations and tables for the white-box
and black-box analyses described in Sections~\ref{sec:blackbox}
and~\ref{sec:whitebox}.

\subsection{Class-wise AD/FD Statistics}
\begin{table}[H]
\centering
\caption{Per-class object deviation statistics. Classes with $N < 15$ are aggregated into \textit{Miscellaneous} (\textit{tennis racket, motorcycle, sports ball, backpack, object, stop sign, handbag, parking meter, suitcase, bench, potted plant, chair}). Q1 and Q3 denote the 25th and 75th percentiles of AD.}
\label{tab:class_deviation}
\small
\begin{tabular}{lrccccc}
\toprule
\textbf{Class} & \textbf{N} & \textbf{Mean AD} & \textbf{Q1 AD} & \textbf{Q3 AD} & \textbf{Max AD} & \textbf{Mean FD} \\
\midrule
Bus                  &    47 & 0.663 & 0.085 & 0.763 & 3.931 & 1.889 \\
Fire hydrant         &    17 & 0.540 & 0.117 & 1.003 & 2.037 & 1.454 \\
Truck                &   183 & 0.505 & 0.089 & 0.703 & 3.548 & 1.419 \\
Traffic light        &   531 & 0.482 & 0.071 & 0.624 & 4.534 & 1.399 \\
Person               &   604 & 0.458 & 0.059 & 0.643 & 5.276 & 1.318 \\
Bicycle              &    28 & 0.399 & 0.059 & 0.565 & 1.888 & 1.022 \\
Car                  &  1566 & 0.348 & 0.054 & 0.393 & 7.782 & 0.984 \\
Umbrella             &    20 & 0.284 & 0.029 & 0.316 & 1.896 & 0.759 \\
\midrule
\textit{Misc.\ ($N<15$)} &    66 & 0.403 & 0.130 & 0.580 & 1.588 & 1.154 \\
\bottomrule
\end{tabular}
\end{table}
% ─────────────────────────────────────────────────────────────────────────────
\subsection{Black-box Scene Visualizer}
\label{app:scene_explorer}
\begin{figure}[H]
    \centering
    \includegraphics[width=\linewidth]{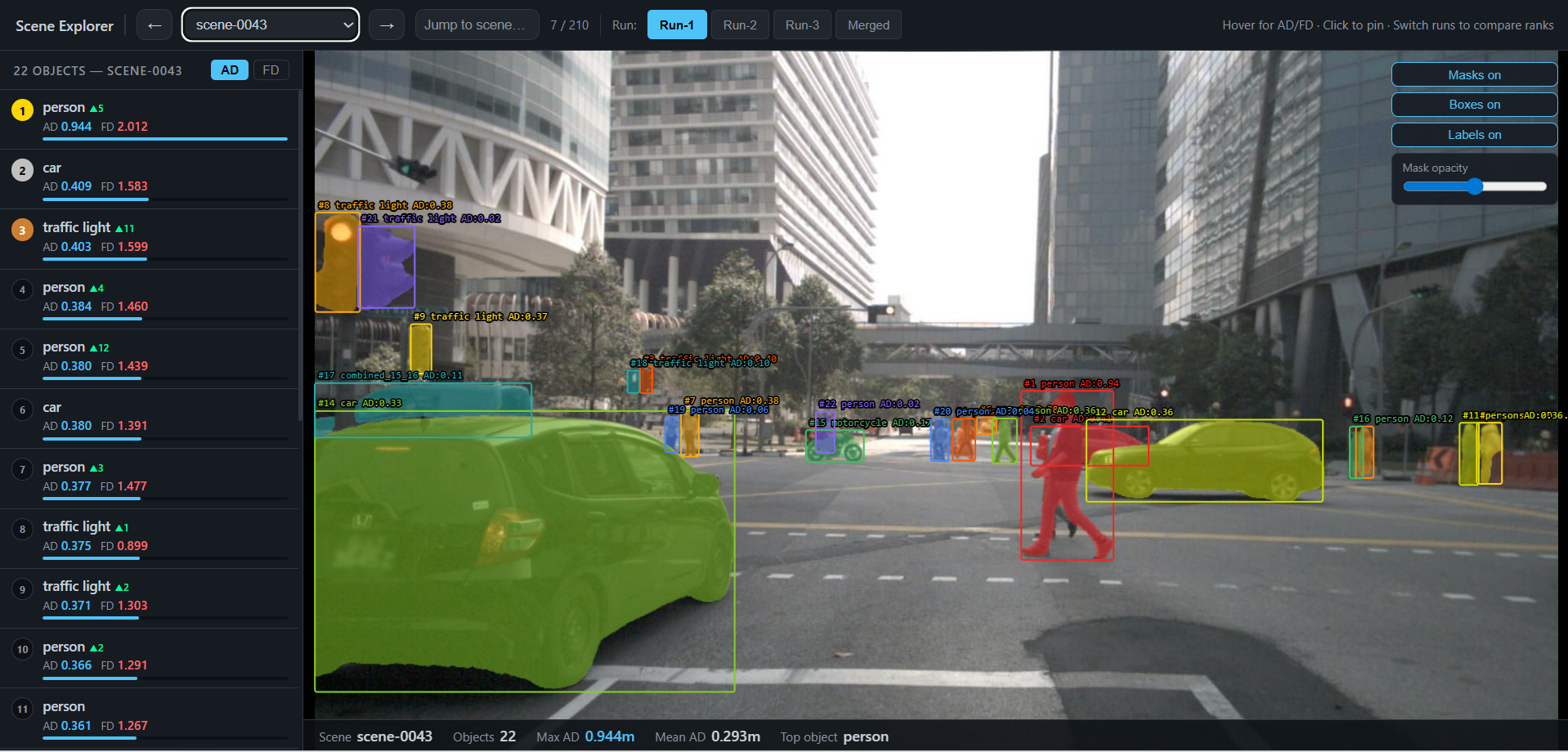}
    \caption{
        Scene Explorer --- the interactive black-box visualisation
        dashboard for Counter\_nuScenes.
        The left panel lists all objects in the scene ranked by AD,
        with gold/silver/bronze chips for the top-3 and colour-coded
        AD/FD bars; arrows indicate rank shift relative to the
        previous run.
        The right panel overlays per-object segmentation masks
         with
        bounding boxes and AD labels.
        The bottom bar shows scene-level statistics (object count,
        max AD, mean AD, top object).
        Multiple runs (Run-1, Run-2, Run-3, Merged) can be selected
        to compare rank orderings across seeds.
        Scene-0043 is shown: a pedestrian (AD $0.944$\,m) ranks
        first despite a large car occupying most of the frame,
        illustrating cases where physical scale does not predict
        causal influence.
    }
    \label{fig:scene_explorer}
\end{figure}

\subsection{Hidden-State Delta Curves}
\label{app:delta_curves}

\begin{figure}[H]
    \centering
    \includegraphics[width=\linewidth]{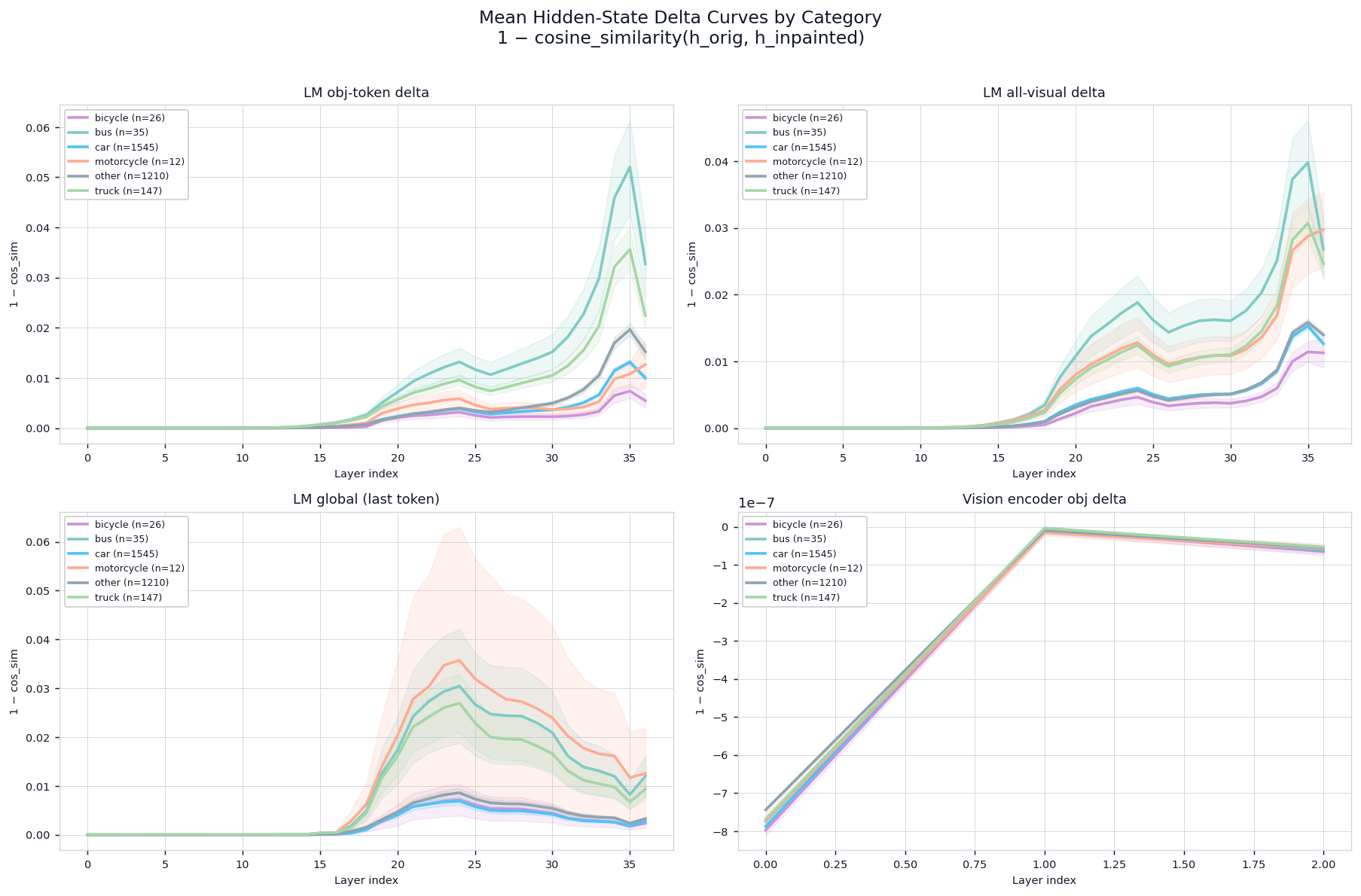}
    \caption{
        Mean hidden-state delta curves ($1 - \cos_{\text{sim}}$,
        original vs.\ inpainted) by object category across all
        LM layers and vision encoder blocks.
        \textbf{Top left}: object-token delta, averaged over
        patch positions spatially covering the removed object.
        \textbf{Top right}: all-visual delta, averaged over all
        180 visual tokens; when this substantially exceeds the
        object-token delta, the removal is rippling into
        surrounding context.
        \textbf{Bottom left}: global delta at the last token
        (scene summary before the expert).
        \textbf{Bottom right}: vision encoder object delta at
        three probed blocks (8, 20, 26), scaled to $10^{-7}$;
        the negative values indicate the inpainted region is
        more self-consistent than the original.
        Shaded bands show $\pm1$ standard error.
    }
    \label{fig:delta_curves}
\end{figure}
\raggedbottom
% ─────────────────────────────────────────────────────────────────────────────
\subsection{Per-Scene Z-Score Distribution}
\label{app:zscore}

\begin{figure}[H]
    \centering
    \includegraphics[width=\linewidth]{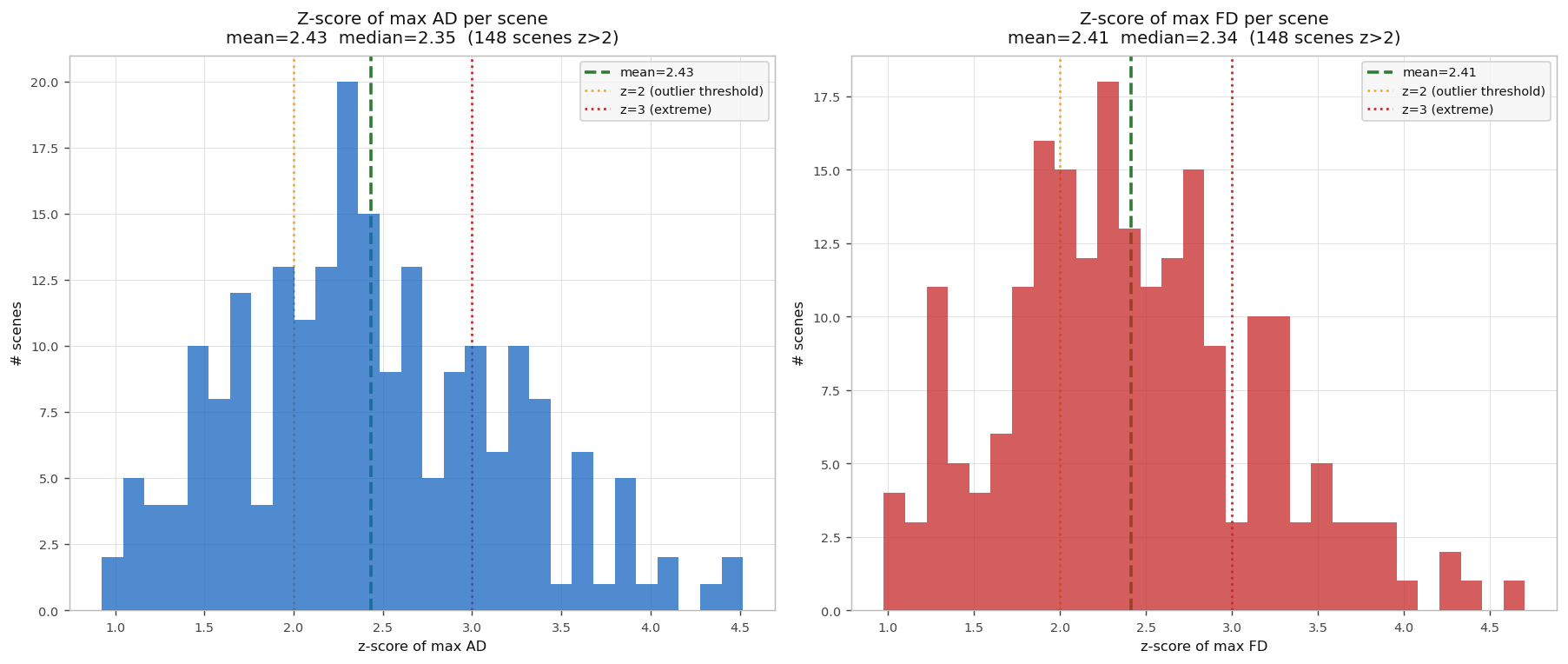}
    \caption{
        Distribution of the z-score of the maximum-AD object per
        scene, $z = (\max(\text{AD}) - \bar{\text{AD}}) /
        \sigma_{\text{AD}}$, for both AD (left) and FD (right).
        Dashed lines mark $z = 2$ (strong outlier) and $z = 3$
        (extreme outlier).
        148 of 210 scenes (70.5\%) exceed $z > 2$ for both
        metrics, indicating widespread single-object dominance
        across the dataset.
        See Section~\ref{sec:blackbox} for discussion.
    }
    \label{fig:zscore}
\end{figure}

% ─────────────────────────────────────────────────────────────────────────────

\end{document}